\documentclass[10pt,twocolumn,letterpaper]{article}

\usepackage{iccv}
\usepackage{times}
\usepackage{subfigure}
\usepackage{epsfig}
\usepackage{graphicx}
\usepackage{algorithm}
\usepackage{algorithmicx}
\usepackage{algpseudocode}
\usepackage{amsmath}
\usepackage{amssymb}
\usepackage{multirow}

\usepackage[breaklinks=true,bookmarks=false]{hyperref}

\iccvfinalcopy 

\def\iccvPaperID{14} 

\begin{document}

\title{Application-Driven AI Paradigm for Person Counting in Various Scenarios}



\author{Minjie Hua\\
China Unicom\\
{\tt\small huamj5@chinaunicom.cn}
\and
Yibing Nan\\
China Unicom\\
{\tt\small nanyb5@chinaunicom.cn}
\and
Shiguo Lian\\
China Unicom\\
{\tt\small sg\_lian@163.com}
}

\maketitle

\begin{abstract}

Person counting is considered as a fundamental task in video surveillance. However, the scenario diversity in practical applications makes it difficult to exploit a single person counting model for general use. Consequently, engineers must preview the video stream and manually specify an appropriate person counting model based on the scenario of camera shot, which is time-consuming, especially for large-scale deployments. In this paper, we propose a person counting paradigm that utilizes a scenario classifier to automatically select a suitable person counting model for each captured frame. First, the input image is passed through the scenario classifier to obtain a scenario label, which is then used to allocate the frame to one of five fine-tuned models for person counting. Additionally, we present five augmentation datasets collected from different scenarios, including side-view, long-shot, top-view, customized and crowd, which are also integrated to form a scenario classification dataset containing 26323 samples. In our comparative experiments, the proposed paradigm achieves better balance than any single model on the integrated dataset, thus its generalization in various scenarios has been proved.
\end{abstract}

\section{Introduction}\label{sec:intro}


Person counting is a common application in many video surveillance tasks such as visitor analysis, traffic monitoring and abnormality recognition. In general, the two mainstream techniques for person counting are detection-based methods and density-based methods. The former detects the whole or part of a person's body and count the number of detected bounding boxes as the final prediction, while the latter generates a density map and sums up all the pixel values to produce the person counting estimation.

Over the past few years, a multitude of algorithms based on convolutional neural networks (CNN) have been developed to handle various real-world scenarios. For most mid-range scenarios where people are captured from a side view, widely-used object detection methods such as YOLO~\cite{YOLO,YOLOv2,YOLOv3,YOLOv4} and SSD~\cite{SSD}, pre-trained on MSCOCO~\cite{MSCOCO}, already satisfy the requirements. However, in some challenging scenarios \eg long-shot images with tiny person targets, top-view images where only heads are visible, or specific places where people are wearing customized suits, additional datasets are necessary for fine-tuning the detector. Furthermore, when facing the scenario of crowd, the accuracy of almost all person detectors drops significantly due to severe body occlusion that impedes feature extraction of individuals. Compared to detection-based approaches, density-based methods~\cite{DensityCount,ShanghaiTech,CSRNet,Bayesian,DMCount,P2PNet} perform better under the crowd scenarios.



Although solutions can be found on a case-by-case basis, experienced engineers must review all video streams in advance and manually assign the most suitable person counting model to each camera device in practical applications. This process is time-consuming and cumbersome, especially as the deployment scale expands. Moreover, when the backstage manager controls the camera by moving, rotating or zooming, it can cause a shift in the captured scene, potentially resulting in imprecise output from the predetermined model.


In this paper, we propose a unified framework that concatenates a scenario classifier and a person counting module. The scenario classifier is a ResNet-50~\cite{ResNet} network, and the person counting module contains five fine-tuned models prepared for corresponding scenarios: (i) side-view, where people are captured from a side view in mid-range, (ii) long-shot, where person bodies appear tiny, (iii) top-view, where people are captured from an overhead camera, (iv) customized, where people are wearing specific suits (we use protective suits as an example, which are commonly observed in healthcare institutions), and (v) crowd, where numerous people appear in the image. For (i)-(iv), we utilized YOLOv5, an improved version of YOLOv4~\cite{YOLOv4}, as the person counting model. For (v), we employ DM-Count~\cite{DMCount} as the person counting model. Additionally, we enhance the robustness of the aforementioned models in specific scenarios by introducing five augmentation datasets. These datasets are also integrated to form a scenario classification dataset used to train the scenario classifier. Further details about these datasets are provided in Section~\ref{subsec:data}.

The main contributions of this work are three-fold:
\begin{itemize}
  \item We propose an application-driven paradigm that automatically selects proper person counting model for the input image based on the scenario classification result.
  \item We introduce five augmentation datasets to enhance models in specific scenarios, together with a classification dataset to train the scenario classifier.
  \item We conduct comparative experiments to demonstrate the generalization of our proposed framework.
\end{itemize}


The rest of the paper is structured as follows: Section~\ref{sec:relat} reviews the related work on person detection and crowd counting. Section~\ref{sec:sys} presents the proposed framework and datasets. In Section~\ref{sec:exp}, we describe the experiments and demonstrate the results. Finally in Section~\ref{sec:concl}, we draw a conclusion.

\section{Related Work}\label{sec:relat}

\subsection{Person Detection}


Traditional person detection methods relied on handcrafted features, \eg deformable part models (DPM) using histograms of oritented gradients (HOG) features~\cite{DPM1,DPM2} and decision forests models using integral channel features (ICF)~\cite{DF1,DF2,DF3}. With the great success of AlexNet~\cite{AlexNet} in ImageNet~\cite{ImageNet} competition, CNN-based methods became mainstream in computer vision tasks like image classification and object detection.


R-CNN~\cite{RCNN} pioneered to introduce CNN into object detection by implementing the two-stage architecture, \ie a proposal stage and a downstream classification stage. Regarding to the intolerable computational complexity for R-CNN to classify each proposal separately, Fast R-CNN~\cite{FastRCNN} optimized the inference time by executing CNN on the whole image at the beginning to extract features shared by all proposals. Faster R-CNN~\cite{FasterRCNN} replaced the external proposal modules in Fast R-CNN with a region proposal network (RPN), integrating the two stages into an end-to-end framework which could be trained jointly.


One-stage detection methods completely abandoned the proposal stage. YOLO~\cite{YOLO} divided the final feature map into $S\times S$ grid cells and predicted the center coordinate, width and height of the object bounding box in each cell. However, this approach had disadvantage in detecting small objects and clustered objects within a single cell. This issue was resolved in YOLOv2~\cite{YOLOv2} by adopting anchor boxes, whose scales and aspect ratios were set a priori by performing k-means clustering~\cite{Kmeans} on the training dataset. YOLOv3~\cite{YOLOv3} introduced residual connections, which was first presented in ResNet~\cite{ResNet}, to construct a deeper darknet-53 network, and proposed feature pyramid networks (FPN) for higher recall on smaller objects. YOLOv4~\cite{YOLOv4} embedded advanced techniques in the backbone, neck and head of original YOLOv3 network, and deployed bag of freebies including mosaic data augmentation, self-adversarial training, and CIoU loss function, together with bag of specials including Mish activation function, cross mini-Batch normalization and drop-block regularization. Just a few weeks later, Jocher~\etal released YOLOv5 implemented on PyTorch~\cite{PyTorch} aiming to further improve accessibility and achieve greater balance between the effectiveness and efficiency. Similar to YOLO serials, SSD~\cite{SSD} also used anchor boxes with a variety of aspect ratios. The difference is that SSD applied anchor boxes on multiple feature maps with different resolutions, thus being able to detect objects of diverse scales.


Person detection can be achieved by training the aforementioned models on datasets taking 'person' as one of the categories, such as MSCOCO~\cite{MSCOCO}, Pascal VOC~\cite{PascalVOC} and CityPersons~\cite{CityPersons}. In these public datasets, the majority of persons were captured from the frontal or side view with an optimal shot range. In recent years, overhead person detection gained significant attention, and models and datasets specifically designed for detecting persons from a top-view perspective were introduced in several studies~\cite{TOPVIEW, TopviewData}. However, there was a lack of open-source datasets customized to detect persons in long-range fields and those wearing protective suits, which motivated us to collect and utilize such datasets in our experiments.

\begin{figure*}[tbp]%
\centering
\includegraphics[width=1\linewidth]{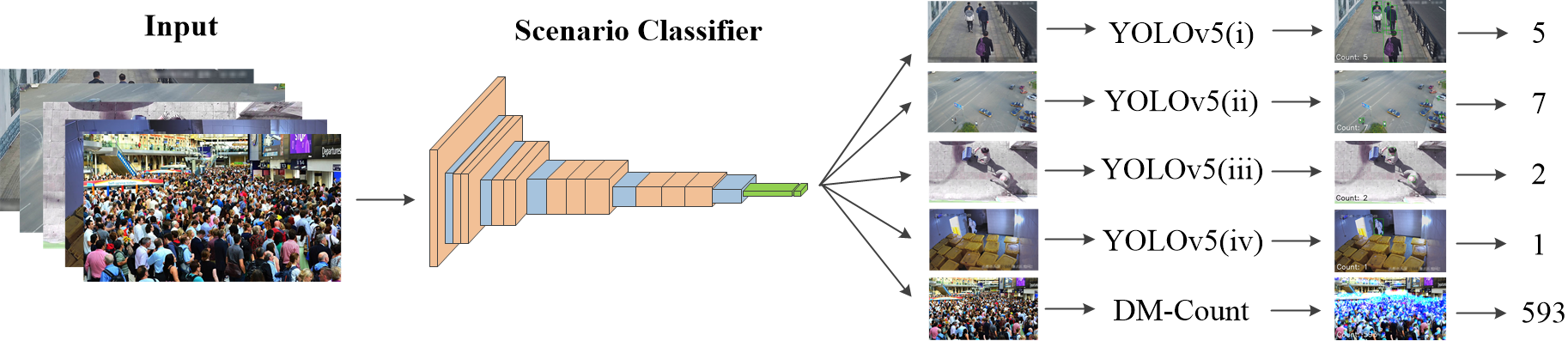}
\caption{The workflow of our proposed paradigm. The input image is first passed to the scenario classifier to obtain a scenario label. Based on this label, the image is allocated to one of five person counting models, which produce the final prediction on the number of persons in the image.}
\label{fig:workflow}
\end{figure*}

\subsection{Crowd Counting}


Early approaches to crowd counting relied on detecting persons, heads or upper bodies~\cite{detCrowd1,detCrowd2,detCrowd3}. However, these detection-based methods suffered from severe occlusions, especially in the dense crowds. Later, researchers developed regression-based frameworks~\cite{regCrowd1,regCrowd2,regCrowd3,regCrowd4} to avoid the detection shortcomings. However, the model was hard to converge without the supervision of head localization annotations in the training process. Recently, methods based on density map estimation~\cite{DensityCount,ShanghaiTech,CSRNet,Bayesian,DMCount,P2PNet} outperformed the aforementioned detection-based and regression-based approaches and became the mainstream solution for crowd counting problem.


Due to the difficulty of delineating the spatial extent for each person in crowd scenes, existing crowd counting datasets~\cite{ShanghaiTech,UCFCC50,UCFQNRF,NWPU} only mark each person with a single dot on the head or forehead. Consequently, the ground truth density map generated from the annotations is a sparse binary matrix, while the predicted density map is a dense real-value matrix. However, directly measuring the discrepancy between the sparse binary and dense real-value matrices with a loss function can make the network hard to converge.



Therefore, a crucial challenge for all density map estimation methods is to effectively utilize the dot annotations. One common approach is to convert each annotated dot into a Gaussian blob, creating a 'pseudo ground truth' that is more balanced. Most prior methods \eg DensityCount~\cite{DensityCount}, MCNN~\cite{ShanghaiTech} and CSRNet~\cite{CSRNet} adopted this idea. However, the kernel widths used for the Gaussian blobs may not accurately reflect the size of people's heads in the image, which can significantly impact the network's performance. Another approach is to design a reasonable loss function. For example, a Bayesian loss was proposed in~\cite{Bayesian} to transform the annotation map into $N$ smoothed density maps, where each pixel value is the posterior probability of the corresponding annotation dot. Recently, the DMCount~\cite{DMCount} model used optimal transport (OT) and total variation (TV) loss to measure the similarity between the normalized predicted density map and the normalized ground truth density map. Without introducing Gaussian smoothing operations, DMCount has been shown to outperform the aforementioned Gaussian-based methods.



\section{Methodology}\label{sec:sys}


\subsection{Overview of the Paradigm}\label{subsec:overview}



The architecture of our proposed paradigm is shown in Fig.~\ref{fig:workflow}. First, the input image is passed to the scenario classifier, which classifies the image into one of five scenario categories as defined in Section~\ref{sec:intro}. After that, the image is fed to the person counting module containing five models fine-tuned on the corresponding augmentation dataset. The person counting module automatically allocate an appropriate model to the image based on its scenario label. Specifically, for (i)-(iv), we apply a fine-tuned YOLOv5 model, named YOLOv5(i) to YOLOv5(iv), to detect and count persons in the image. For (v), we adopt the DM-Count model to estimate the density map, and output the final count prediction by summing up all the pixel values in the map.



\subsection{Scenario Classifier}\label{subsec:scenecls}


The scenario classifier is a fundamental ResNet-50~\cite{ResNet} network. The input layer accepts an image with size of $224\times224\times3$, followed by a $7\times7\times64$ convolution layer, a $3\times3$ max pooling layer and four groups of bottleneck residual blocks. The first block contains three repeated units composed by three convolution layers with kernel size of $1\times1\times64$, $3\times3\times64$ and $1\times1\times256$. The second block contains four repeated units composed by three convolution layers with kernel size of $1\times1\times128$, $3\times3\times128$ and $1\times1\times512$. The third block contains six repeated units composed by three convolution layers with kernel size of $1\times1\times256$, $3\times3\times256$ and $1\times1\times1024$. The fourth block contains three repeated units composed by three convolution layers with kernel size of $1\times1\times512$, $3\times3\times512$ and $1\times1\times2048$. After all the convolution layers, an average pooling layer, a 5d fully connection layer and a softmax layer are applied to predict the scenario category.

\subsection{Person Counting Module}\label{subsec:pcm}


The person counting module includes five models that are automatically executed based on the scenario label of the input image. These models are fine-tuned YOLOv5 and DM-Count models, each enhanced for a specific scenario.

Four YOLOv5 models, referred to as YOLOv5(i) to YOLOv5(iv), are used in scenarios (i)-(iv) to detect and count the number of persons based on their bodies or heads. The YOLOv5 model consists of a backbone to aggregate and form image features at different granularity levels, a neck to mix and combine image features, and a head to consume features from the neck and predict the objectness score, class probability, and bounding box coordinates for each anchor box at multiple scales and aspect ratios. The anchor boxes are pre-defined shapes and sizes that cover different parts of the image.


For scenario (v), the DM-Count is used to estimate the density map for person counting. It adopts VGG-19~\cite{VGG} as the backbone network and employs OT and TV loss instead of traditional Gaussian smoothing operations to avoid hurting the realness of the ground truth. Specifically, the OT loss measures the similarity between the predicted and ground truth density maps, and the TV loss is added to further enhance the smoothness of the predicted density map.

\section{Experiments}\label{sec:exp}

\subsection{Dataset Overview}\label{subsec:data}


To train and validate our five person counting models and scenario classifier, we constructed five augmentation datasets, which were also integrated to form the scenario classification dataset. Each augmentation dataset was collected from a specific scenario as defined in Section~\ref{sec:intro}. The side-view dataset was mainly collected from public streets, parks, and offices \etc, captured by cameras at a height of 3-5 meters, with each person annotated with a body bounding box. The long-shot dataset was collected from cameras placed at far distances from the subjects, such as surveillance cameras on highways, ferries, and squares \etc, with each person also annotated with a body bounding box. The top-view dataset was collected from various sources, including a public overhead person detection dataset~\cite{TopviewData}, with the annotations being head bounding boxes. The protective suit dataset, as an example of customized scenario, was mostly collected from hospitals, health centers, and medical waste rooms, with the dataset characterized by data augmentation on people wearing protective suits and annotated with body bounding boxes. Lastly, the crowd counting dataset was randomly selected from ShanghaiTech~\cite{ShanghaiTech}, UCF-CC50~\cite{UCFCC50}, UCF-QNRF~\cite{UCFQNRF} and NWPU~\cite{NWPU}. with each person marked with a dot on the head. For the scenario classification dataset, we combined all the images from the five augmentation datasets and labeled each image with a tag indicating the corresponding scenario. Additionally, we inferred the number of persons in each image from the raw annotations, and calculated the maximum, minimum and average values for each augmentation dataset and the integrated dataset. The statistics are presented in Table~\ref{tb:statis}.

\begin{table}[tbp]
\center
  \caption{The statistics of five augmentation datasets and the integrated dataset. The columns from left to right show the dataset name, data scale, and the maximum, minimum, and average number of persons respectively.}
  \label{tb:statis}
\begin{center}
\begin{tabular}{c|c|c|c|c}
\hline
Dataset & Scale & Max. & Min. & Avg.\\
\hline
Side-View & 5648 & 70 & 0 & 6.6\\
Long-Shot & 5305 & 16 & 0 & 2.7\\
Top-View & 4632 & 18 & 0 & 3.3\\
Protective-Suit & 5904 & 6 & 0 & 1.8\\
Crowd & 4834 & 12865 & 34 & 743.8\\
Integrated & 26323 & 12865 & 0 & 138.1\\
\hline
\end{tabular}
\end{center}
\end{table}


To train the person counting models, each augmentation dataset is randomly divided into a training set and a validation set at a ratio of 8:2. As for the training of the scenario classifier, the division of the scenario classification dataset keeps consistent with that of the augmentation datasets for the convenience of joint network evaluations, which will be explained in Section~\ref{subsec:eval}.

\begin{table}[tp]
\center
  \caption{The evaluation of the scenario classifier. The one-vs-rest strategy is used to calculate precision, recall, F1-score and support for each class, as well as the macro-average across all classes.}
  \label{tb:cls_eval}
\begin{center}
\begin{tabular}{c|c|c|c|c}
\hline
Class & Prec. & Rec. & F1-score & Support\\
\hline
Side-View & 0.70 & 0.75 & 0.72 & 1130\\
Long-Shot & 0.87 & 0.82 & 0.85 & 1061\\
Top-View & 0.93 & 0.98 & 0.95 & 926\\
Protective-Suit & 0.73 & 0.70 & 0.71 & 1181\\
Crowd & 0.87 & 0.86 & 0.87 & 967\\
Macro-Average & 0.81 & 0.81 & 0.81 & 5265\\
\hline
\end{tabular}
\end{center}
\end{table}

\begin{figure}[tp]%
\centering
\includegraphics[width=1\linewidth]{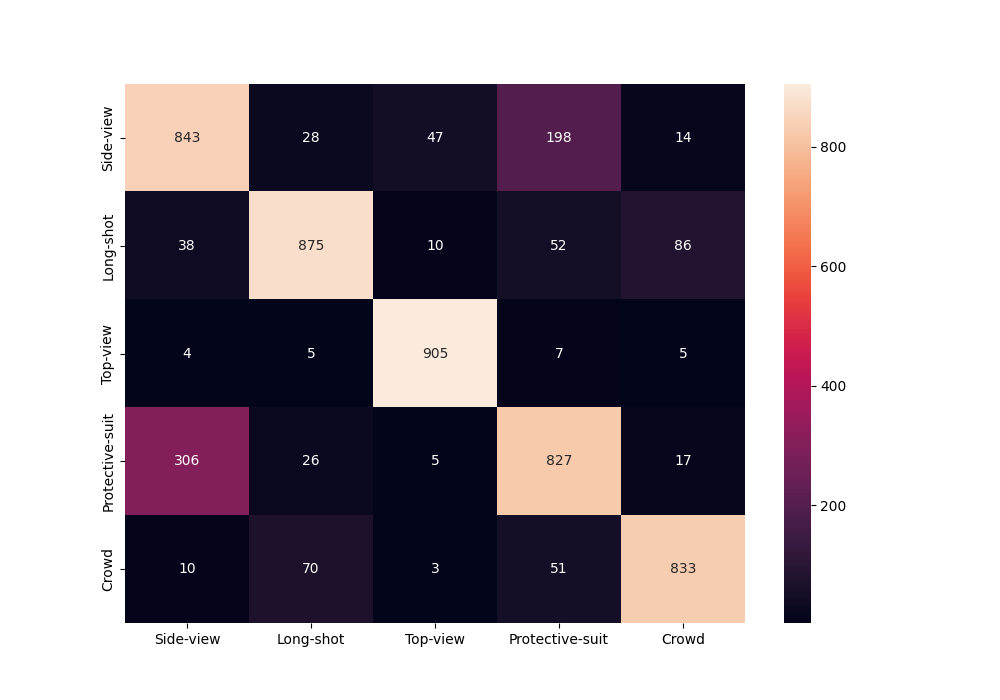}
\caption{The confusion matrix of scenario classifier evaluation. Each grid in the matrix corresponds to a combination of ground truth and prediction, with the horizontal and vertical axes representing the ground truth and prediction respectively.}
\label{fig:cm}
\end{figure}

\subsection{Experiments Setup}

\begin{table*}[htbp]
\center
  \caption{Comparisons on the networks with and without the scenario classifier. The first five rows report the results of fixed person counting models without using the scenario classifier, while the last row shows the performance of the model automatically selected by the scenario classifier. The experiments are conducted on five separate datasets and the integrated dataset, using MAE and RMSE as evaluation metrics.}
  \label{tb:cross_eval}
\begin{center}
\begin{tabular}{c|c|c|c|c|c|c|c|c|c|c|c|c}
\hline
\multirow{2}{*}{Model} & \multicolumn{2}{c|}{Side-View} & \multicolumn{2}{c|}{Long-Shot} & \multicolumn{2}{c|}{Top-View} & \multicolumn{2}{c|}{Protective-Suit} & \multicolumn{2}{c|}{Crowd} & \multicolumn{2}{c}{Integrated}\\  \cline{2-13}
                       & MAE & RMSE & MAE & RMSE & MAE & RMSE & MAE & RMSE & MAE & RMSE & MAE & RMSE\\
\hline
YOLOv5(i) & \textbf{0.39} & \textbf{0.94} & 1.02 & 1.44 & 1.24 & 1.57 & 0.76 & 1.02 & 613.6 & 721.3 & 113.4 & 309.1\\
YOLOv5(ii) & 0.68 & 1.13 & \textbf{0.36} & \textbf{0.80} & 1.21 & 1.48 & 0.91 & 1.12 & 583.0 & 622.5 & 107.7 & 266.8\\
YOLOv5(iii) & 1.88 & 2.31 & 1.38 & 1.61 & \textbf{0.44} & \textbf{0.89} & 1.42 & 1.69 & 624.7 & 736.4 & 115.8 & 315.6\\
YOLOv5(iv) & 0.44 & 0.91 & 1.09 & 1.49 & 1.26 & 1.57 & \textbf{0.28} & \textbf{0.69} & 616.2 & 725.8 & 113.8 & 311.1\\
DM-Count & 7.56 & 10.37 & 13.23 & 18.63 & 6.45 & 8.77 & 6.84 & 10.1 & \textbf{96.4} & \textbf{153.8} & 24.7 & \textbf{66.9}\\
\textbf{Automatic} & 0.41 & 0.93 & 0.58 & 1.01 & 0.66 & 1.23 & 0.54 & 0.92 & 108.6 & 168.1 & \textbf{20.4} & 72.0\\
\hline
\end{tabular}
\end{center}
\end{table*}


The experiment projects are implemented on a workstation equipped with double Nvidia 3080Ti GPUs. The scenario classifier and five person counting models are trained separately. For the scenario classifier, we utilize cross-entropy as the loss function and Adam~\cite{ADAM} as the optimizer, with a learning rate of 0.001. The person counting module is built on YOLOv5 and DM-Count. During the training phase of YOLOv5, the loss function is a combination of bounding box regression loss, objectness loss and classification loss, and is optimized using stochastic gradient descent (SGD) with an initial learning rate of 0.01 and momentum of 0.937. To train DM-Count, we use Adam with a learning rate of $1\times 10^{-5}$ to optimize the overall loss function, which combined counting loss, OT loss and TV loss.

During the inference phase of YOLOv5, all the generated proposals are first filtered by non-maximum suppression (NMS) with intersection over union (IoU) threshold of 0.45 and confidence threshold of 0.6. The model then outputs the number of reserved bounding boxes as the final prediction. For DM-Count, the model produces a density map, and the sum of all the pixel values in the map is rounded to estimate the number of persons in the image.

\subsection{Evaluations}\label{subsec:eval}

The evaluations are designed to consider two aspects: the performance of the trained scenario classifier, and the effect of the scenario classifier on the overall network.

To evaluate the performance of our trained scenario classifier, we input all 5265 validation samples into the model and analyze the predictions. We adopt the one-vs-rest strategy to convert our multi-class problem into a series of binary ones. For each class, all the remaining classes are treated as the negative class, enabling us to use binary classification metrics such as precision, recall, and F1-score in a multi-class problem. Given the relatively balanced class distribution in our dataset, we use the macro average (the arithmetic mean of all metrics across classes) to reflect the performance of the scenario classifier on the entire validation dataset. The evaluation result and the confusion matrix are presented in Table~\ref{tb:cls_eval} and Fig.~\ref{fig:cm}. It is observed that the scenario classifier produces generally fair predictions. However, some scenarios, such as side-view and protective-suit, as well as long-shot and crowd, are more frequently mislabeled.

Regarding the effect of the scenario classifier, we compare the networks with and without the scenario classifier. When using the scenario classifier, the person counting model is automatically selected based on the scenario classification result. Conversely, when not using the scenario classifier, the person counting model is fixed to one of the five person counting models. The evaluations are performed on the five augmentation datasets and the integrated dataset. We adopt MAE and RMSE to measure the difference between ground truth and the model's prediction. Recall that MAE and RMSE are calculated as follows:

\begin{equation}\label{eq:metrics}
\begin{aligned}
\begin{array}{l}
    {\text{MAE}} = \frac{1}{n}\sum_{i=1}^{n}\left | {y}_{i}-\hat{y}_{i}  \right |\\
    \\
    {\text{RMSE}} = \sqrt{\frac{1}{n}\sum_{i=1}^{n} \left ( {y}_{i}-\hat{y}_{i} \right )^{2} }
\end{array}
\end{aligned}
\end{equation}
where $n$ is the number of validation samples, ${y}_{i}$ and $\hat{y}_{i}$ are the ground truth and prediction of the $i$-th sample respectively. As shown in Table~\ref{tb:cross_eval}, each person counting model performs best on a specific scenario but produces higher deviation on other datasets. In comparison, with automatic model selection by the scenario classifier, our network achieves a great balance on the integrated dataset.

\begin{figure*}[tbp]%
\centering
\includegraphics[width=1\linewidth]{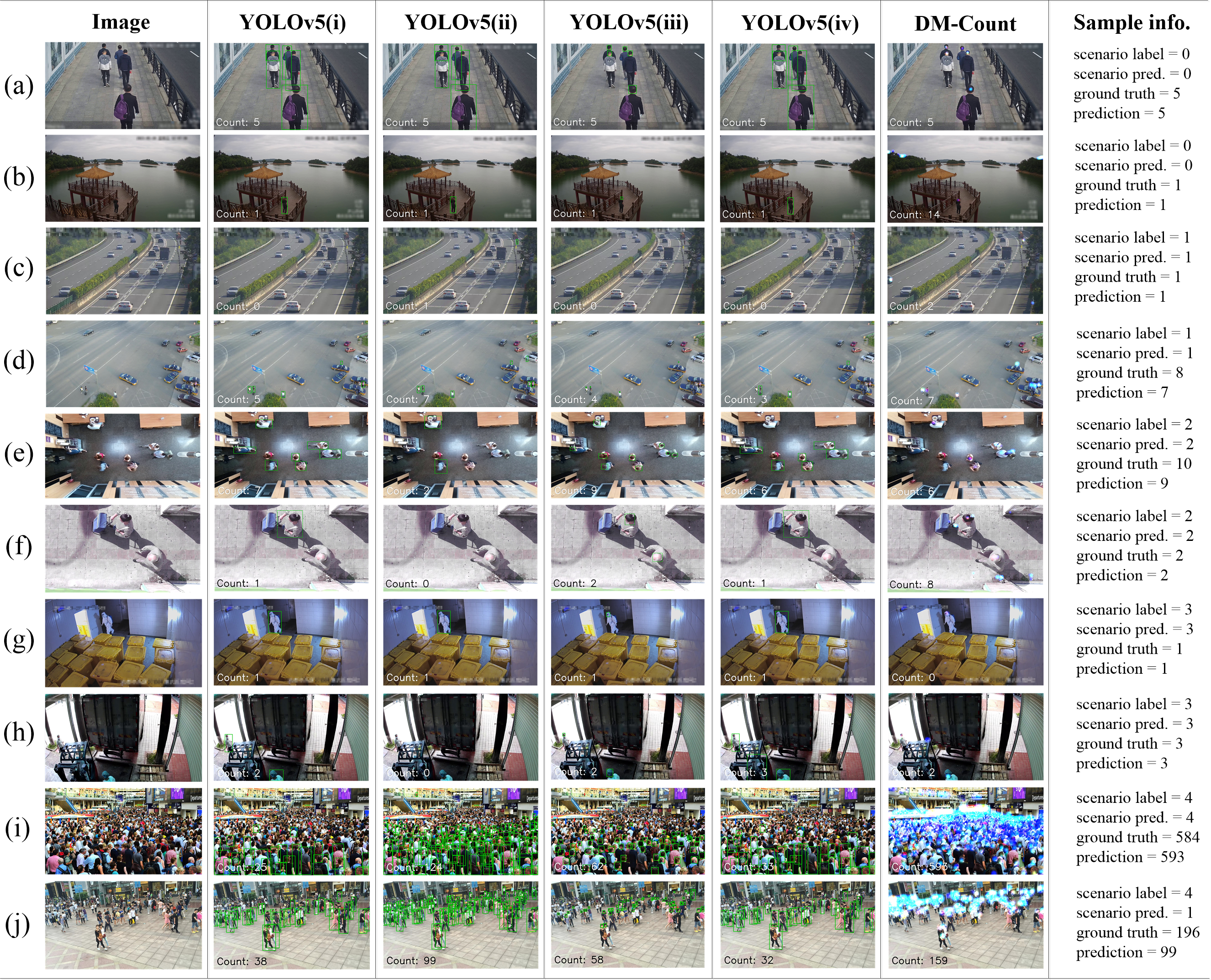}
\caption{Visualization results of our experiments. The input images are picked from five augmentation datasets and each image is processed by all the five person counting models. For YOLOv5 models, we render the detected body/head bounding boxes on the image. For DM-Count model, we overlay the generated heatmap on the original image. And the counting result is remarked on each output image. The last column indicates the actual and the predicted scenario labels of the image, along with the ground truth and final prediction of person counting for each image. In case (a)-(i), the scenario classifier successfully allocates the input image to the correct person counting model, while in case (j), the input image is mislabeled thus passed to a model that is not the most appropriate.}
\label{fig:result}
\end{figure*}

\subsection{Visualization Results}


As illustrated in Fig.~\ref{fig:result}, we visualize the inference results of person counting models on several samples to demonstrate the performance of our proposed paradigm. Specifically, we select two validation images from each augmentation dataset and apply all the person counting models for prediction. For YOLOv5(i), (ii) and (iv), we render body bounding boxes on the image, while for YOLOv5(iii), we render head bounding boxes. For the DM-Count model, we first blur the estimated density map with a Gaussian kernel with size of $5\times5$ to generate a heatmap, which is then overlaid onto the original image for better visualization. We also remark the counting number on the left-bottom of each output image. From these results, we observe that each person counting model shows its strengths and weaknesses in specific scenarios. Moreover, the existence of scenario classifier raises the possibility for the input image to be processed by the appropriate person counting model.




\section{Conclusion}\label{sec:concl}


In this work, we propose an AI paradigm specially designed for person counting task, which takes into account the scenario in which the image is captured. The proposed architecture consists of a scenario classifier and a person counting module containing four YOLOv5 models and a DM-Count model, each fine-tuned for a specific scenario. The scenario classifier is proved effective in allocating the input image to one of five person counting models based on its scenario label. The YOLOv5-based models count the persons by detecting bodies or heads and counting the number of bounding boxes, while the DM-Count produces an estimation by generating a density map and summing up all the pixel values. Our paradigm outperforms any single predetermined model on the integrated validation dataset, demonstrating its generalization in various scenarios.

{\small
\bibliographystyle{ieee}
\bibliography{egbib}
}

\end{document}